# Reducing Uncertainty in Navigation and Exploration

Kenneth Basye[*]    Moises Lejter[†]    Keiji Kanazawa[‡]

Department of Computer Science
Brown University
Box 1910, Providence, RI 02912

## Abstract

A significant problem in designing mobile robot control systems involves coping with the uncertainty that arises in moving about in an unknown or partially unknown environment and relying on noisy or ambiguous sensor data to acquire knowledge about that environment. We describe a control system that chooses what activity to engage in next on the basis of expectations about how the information returned as a result of a given activity will improve its knowledge about the spatial layout of its environment. Certain of the higher-level components of the control system are specified in terms of probabilistic decision models whose output is used to mediate the behavior of lower-level control components responsible for movement and sensing.

## 1   Introduction

We are interested in building systems that construct and maintain representations of their environment for tasks involving navigation. Such systems should expend effort on the construction and maintenance of these representations commensurate with expectations about their value for immediate and anticipated tasks. Such systems should employ expectations about the accuracy of the information returned from sensors to assist in choosing activities that are most likely to improve the accuracy of its representations. Finally, such systems should employ expectations about the time required to carry out a given activity in order to make appropriate tradeoffs regarding other activities whose expected return value depends on when they are completed or how long they take.

---
[*]kjb@cs.brown.edu
[†]mlm@cs.brown.edu
[‡]kgk@cs.brown.edu

An architecture for the above sort of systems must be able to generate hypotheses, reason about the impact of actions designed to confirm or refute those hypotheses, and revise its hypotheses in the light of new information. In the remainder of this paper, we will describe a particular robot control system, but the discussion focuses on the issues and the specific decision making technologies that we have chosen to address those issues.

## 2   Exploration and Navigation

We start with the premise that having a map of your environment is generally a good thing if you need to move between specific places whose locations are clearly indicated on that map. The more frequent your need to move between locations, the more useful you will probably find a good map. If you are not supplied with a map and you find yourself spending an inordinate amount of time blundering about, it might occur to you to build one, but the amount of time you spend in building a map will probably depend upon how much you anticipate using it. Once you have decided to build a map, you will have to decide when and exactly how to go about building it. Suppose that you are on an errand to deliver a package and you know of two possible routes, one of which is guaranteed to take you to your destination and a second which is not. By trying the second route, you may learn something new about your environment that may turn out to be useful later, but you may also delay the completion of your errand.

Suppose you come to an intersection that looks familiar; the intersection that you recall had a store belonging to a chain of department stores on one of the corners. Establishing that there is indeed a store from the same chain on one of the corners of the the present intersection is not enough to determine that the present and recalled intersections are the same, but it will certainly strengthen the hypothesis that they are the same. There might be other information that you could gather in an effort to confirm or deny the hypothesis. For instance you might recall that two blocks west of the recalled intersection there is a certain restaurant. On the basis of how much time you have and how important it is for you to establish your location with regard to that recalled intersection, you will have to choose whether or not it is worth your while to look for the restaurant; such information could be very costly if you're not sure which direction is west or there is danger of getting lost in a maze of one-way streets.

Our robot, Huey, has strategies for checking out many simple geometric features found in typical office environments; we refer to these strategies as *feature detectors*. The complete set of feature detectors used by Huey and the details concerning their implementation are described in [Randazza, 1989]. Each feature detector is realized as a control process that directs the robot's movement and sensing. On the basis of the data gathered during the execution of a given feature detector, a probability distribution is determined for the random variable corresponding to the proposition that the feature is present at a specific location.

Huey is designed to explore its environment in order to build up a representation of that environment suitable for route planning. In the course of exploration, Huey induces a graph that captures certain qualitative features of its environment [Kuipers and Byun, 1988, Levitt *et al.*, 1987, Basye *et al.*, 1989]. In addition to detecting geometric features like corners and door jambs, Huey is able to classify locations. In particular, Huey is able to distinguish between corridors and places where corridors meet or are punctuated by doors leading to offices, labs, and storerooms. A corridor is defined as a piece of rectangular space bounded on two sides by uninterrupted parallel surfaces 1.5 to 2 meters apart and bounded on the other two sides by *ports* indicated by abrupt changes in one of the two parallel surfaces. The ports signal *locally distinctive places* (LDPs) (after [Kuipers, 1978]) which generally correspond to hallway junctions. Uninterrupted corridors are represented as arcs in the induced graph while junctions are represented as vertices. Junctions are further partitioned into classes of junctions (*e.g.,* L-shaped junctions where two corridors meet at right angles, or T-shaped junctions where one corridor is interrupted by a second perpendicular corridor). Huey is given a set of junction classes that it uses to classify the locations encountered during exploration, and label the vertices in its induced graph to support route planning.

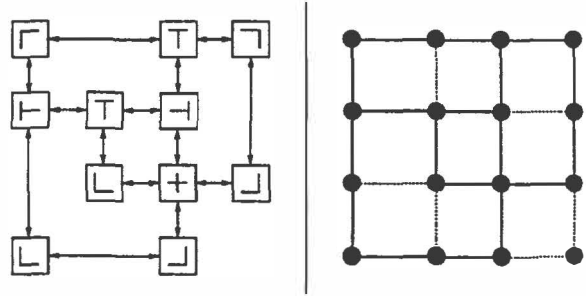

Figure 1: A map and the underlying grid

# 3 Value of Exploration

There are many ways to design a decision model for allocating time to exploration and errand running. Our treatment here is meant to provide an example of the sort of decision models used in Huey.

For the simple model presented here, we assume that the system of junctions and corridors that make up Huey's environment can be registered on a grid so that every corridor is aligned with a grid line and every junction is coincident with the intersection of two grid lines. In the following, the set of junction types, $J$, corresponds to all possible configurations of corridors incident on the intersection of two grid lines. Intersections with at least one incident corridor correspond to LDPs. Since we also assume that Huey knows the dimensions of the grid (*i.e.,* the number of $x$ and $y$ grid lines), Huey can enumerate the set of possible maps $M = \{M_1, M_2, \ldots, M_m\}$, where a map corresponds to an assignment of a junction type to each intersection of grid lines. For most purposes, we can think of a map as a labeled graph. Figure 1 shows a junction assignment and the associated grid.

We restrict $M$ by making an assumption about office buildings of the sort that Huey will find itself in, that all the LDPs are connected. However, for most situations, this restriction does not reduce the size of $M$ sufficiently; in the next section we discuss several strategies for restricting $M$ to a manageable size. We now turn to the details of the decision model used for exploration.

In the simple model considered here, we assume that some paths for completing a task are known, but they may all be longer than necessary. Huey has to choose between taking the shortest path through known territory, and trying the shortest path consistent with what is known. In the latter case, Huey will learn something new, but it may end up taking

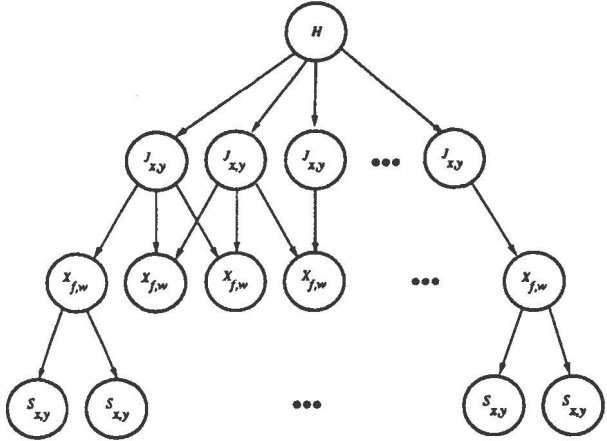

Figure 2: The probabilistic model used for reasoning about tradeoffs

longer to complete its task.

Let $L$ be the set of all locally distinctive places in the robot's environment, $C = \{C_1, C_2, \ldots, C_n\}$ be a set of equivalence classes that partitions $L$, and $F$ be a set of primitive geometric features (*e.g.*, convex and concave corners, flat walls). Each class in $C$ can be characterized as a set of features in $F$ that stand in some spatial relationship to one another. As Huey exits a port, a local coordinate system is set up with its origin on the imaginary line defined by the exit port and centered in the corridor. The space about the origin enclosing the LDP is divided into a set of equi-angular wedges $W$. For each feature/wedge pair $(f, w)$ in $F \times W$, we define a specialized feature detector $d_{f,w}$ that is used to determine if the current LDP satisfies the feature $f$ at location $w$ in the coordinate system established upon entering the LDP, and a boolean variable of the form, $X_{f,w}$, used to represent whether or not the feature $f$ is present at location $w$.

Let $H$ be a random variable corresponding to the actual configuration of the environment; $H$ takes on values from $M$. Let $J_{x,y}$ be a random variable corresponding to the junction type of the intersection at the coordinates, $\langle x, y \rangle$, in the grid; $J_{x,y}$ can take on values from the set $C$. Let $X_{f,w}$ be as above, a boolean variable corresponding to the presence of a feature at a particular position. Let $S_{x,y}$ be a random variable corresponding to a possible sensing action taken at the coordinates, $\langle x, y \rangle$, in the grid. Let $S$ correspond to the set of sensing actions taken thus far. The complete probabilistic model is shown in Figure 2.

In our simple model, Huey has to decide between the two alternatives, $P_K$ and $P_U$, corresponding to paths through known and unknown territory. To compute $\mathrm{Bel}(H) = \mathrm{Pr}(H|S)$, $\mathrm{Pr}(H)$ is assumed to be uniform, $\mathrm{Pr}(J_{x,y}|H)$ and $\mathrm{Pr}(X_{f,w}|J_{x,y})$ are determined by the geometry, and $\mathrm{Pr}(S_{x,y}|X_{f,w})$ is determined experimentally. Let $T = \{T_1, T_2, \ldots, T_r\}$ denote the set of all tasks corresponding to point-to-point traversals, and $\mathrm{E}(|T_i|)$ denote the expected number of tasks of type $T_i$. Let $\mathrm{Cost}(T_i, M_j, M_k)$ be the time required for the task $T_i$ using the map $M_j$, given that the actual configuration of the environment is $M_k$; if $M_j$ is a subgraph of $M_k$, then $\mathrm{Cost}(T_i, M_j, M_k)$ is just the length of the shortest path in $M_j$. For evaluation purposes, Huey assumes that it will take at most one additional exploratory step. Let $T^*$ denote Huey's current task, and $M^*$ the map currently used for route planning. In the simplest model, Huey correctly classifies any location it passes through, and $M^*$ is the minimal assignment consistent with what it has classified so far.

To complete the decision model, we need a means of computing the expected value of $P_K$ and $P_U$. In general, the value of a given action is the sum of the immediate costs related to $T^*$ and the costs for expected future tasks. Let

$$\mathrm{Futures}(M_i, I) = \sum_{j=1}^{r} \mathrm{E}(|T_j|)\,\mathrm{Cost}(T_j, M_I^*, M_i),$$

where $M_I^* = M_{\arg\max_j \mathrm{Pr}(M_j|I)}$.

If classification is perfect, Huey correctly classifies any location it passes through, and $M_{\mathcal{E}}^*$ is the minimal assignment consistent with what it has classified so far. In this case, the expected value of $P_K$ is

$$\mathrm{Cost}(T^*, M^*, \_) + \mathrm{Futures}(\_, \mathcal{E}).$$

If classification is imperfect, the expected value of $P_K$ is

$$\sum_{j=1}^{m} \mathrm{Pr}(M_j|\mathcal{E})\left[\mathrm{Cost}(T^*, M^*, M_j) + \mathrm{Futures}(M_j, \mathcal{E})\right].$$

Handling $P_U$ is just a bit more complicated. Suppose that Huey is contemplating exactly one sensing action that will result in one of several possible observations $O_1, \ldots, O_n$, then the expected value of $P_U$ is

$$\sum_{j=1}^{m} \mathrm{Pr}(M_j|\mathcal{E})\,\mathrm{Cost}(T^{*'}, M^*, M_j) +$$
$$\sum_{i=1}^{n} \mathrm{Pr}(O_i) \sum_{j=1}^{m} \mathrm{Pr}(M_j|O_i, \mathcal{E})\mathrm{Futures}(M_j, [O_i, \mathcal{E}])$$

where $T^{*'}$ is a modification of $T^*$ that accounts for the proposed exploratory sensing action.

We use Jensen's variation of Lauritzen and Spiegelhalter's algorithm [Jensen, 1989, Lauritzen and Spiegelhalter, 1988] to compute the belief function, and compute $\sum_{k=1}^{r} E(|T_k|) \text{Cost}(T_k, M_i, M_j)$ once for all pairs $\langle M_i, M_j \rangle \in M \times M$, storing the results in a table.

The time required to compute the belief function is determined by the size of the sample spaces for the individual random variables and the connectivity of the network used to specify the decision model. In the case of a singly-connected[1] network, the cost of computation is polynomial in the number of nodes and the size of the largest sample space—generally the space of possible maps. The network shown in Figure 2 would be singly-connected if each feature, $X_{f,w}$, had at most one parent corresponding to a junction, $J_{x,y}$. In the case of a multiply-connected network, the cost of computation is a function of the product of the sizes of the sample spaces for the nodes in the largest clique of the graph formed by triangulating the DAG corresponding to the original network. In this case, the multiply-connected network is more appropriate than the singly-connected one because the presence of a feature at one junction can affect the classification of a neighboring junction. Because these networks have very large cliques, they can require long computations, or may even exceed space limitations.

In the singly-connected case, the only stumbling block is the size of the hypothesis node. We can get around this problem by limiting the number of alternative hypothesis actually considered at any given time. In this case the space of possible maps chosen may not include the map corresponding to the actual configuration of the environment. To detect this occurance, we include in the state space of the hypothesis node a value which represents the probability that none of the other maps is correct; we refer to this as the *none of the above, or NOTA* state. The conditional probabilities for any junction node given this state are uniform over all the possibilities for the junction. That is, the NOTA state provides no information about any junction. Adding evidence which is consistent with one or more maps in the hypothesis node causes the posterior probability of the NOTA state to fall. Only when evidence is added which conflicts with *all* maps in $H$ can the NOTA state become more probable. If the value of the NOTA state exceeds the values of all the other states in $H$, we dynamically adjust the model by generating a new set of maps to replace the old set. The new set is chosen to be as consistent as possible with the current evidence. If enough evidence has been collected, the new set may be exhaustive, otherwise the procedure is repeated until an exhaustive set is possible. In the non-exhaustive cases, including the initial case, the hypotheses are generated randomly, but not uniformly. All of the hypotheses will be completely connected, as discussed above. In addition, we use the ratio of the number of edges in the hypothesis to the number of total possible edges as a measure of density, and we prefer to generate maps of medium density over very sparse or very dense maps.

| Size of $H$ | Length of Exploration | Time (seconds) | Cost of Largest Clique |
|---|---|---|---|
| 10 | 4 | 92.2 | 88064 |
| 10 | 6 | 35.6 | 41728 |
| 10 | 8 | 8.8 | 7424 |
| 10 | 10 | 14.6 | 12992 |
| 20 | 4 | 148.0 | 147456 |
| 20 | 6 | 73.6 | 73728 |
| 20 | 8 | 28.9 | 26240 |
| 20 | 10 | 14.3 | 13152 |
| 30 | 4 | 238.3 | 245760 |
| 30 | 6 | 129.7 | 150528 |
| 30 | 8 | 65.4 | 64512 |
| 30 | 10 | 12.8 | 10944 |

Table 1: Average results for 10 runs on a $4 \times 4$ grid.

In the case of multiply-connected networks, an additional source of complexity arises in the connections between the junctions and the feature detectors. Any large, unexplored block of space will result in a large block of multiply connected junction nodes, which will in turn in a large clique after triangulation. Table 1 shows how exploration and the size of the hypothesis node affect the time to update the network and the product of the state spaces for the largest clique. By exploring only a little more, this critical product can be reduced substantially. The next section outlines an approach that allows Huey to reduce the connectivity of the network used to encode the decision model through the use of a hierarchy of ever-more-detailed networks.

[1] A network is said to be singly-connected if there is at most one directed path between any two nodes; otherwise, it is said to be multiply connected [Pearl, 1988].

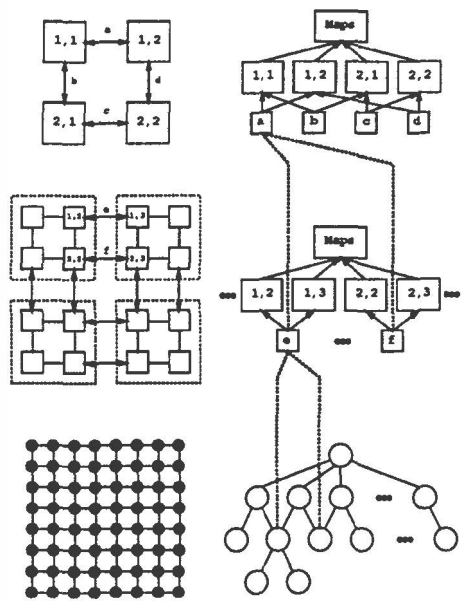

Figure 3: Grid Hierarchy

## 4 Decision Models for Early Exploration

The approach outlined in Section 3 attempts to reason about the best alternative path to follow at any one time based on an analysis of the information known about the whole map. Although this approach allows us to compute optimal (or near-optimal) decisions, it has some drawbacks. In particular, if too little information about the world is known when we attempt to evaluate the decision network that algorithm builds, the number of possibilities that must be taken into account blows up exponentially, making the evaluation impossible. The larger the size of the world the robot finds itself in, the more likely this situation will be. Two simple strategies exist to deal with this problem: one is to guarantee that enough information will be available by the time the robot attempts to evaluate the decision network described above; the other, to try and limit the size of the network our algorithm will have to analyze. In [Dean et al., 1990], we outline a solution to this problem based only on the first of these two strategies. A better solution, which employs both strategies, is outlined here.

Our solution consists of analyzing the robot's current situation using a hierarchy of decision networks of increasing degree of abstraction, as illustrated in Figure 3. The bottom, most detailed, network in the hierarchy is the one corresponding to an accurate representation of the real world. It is the same as the decision network described in Section 3. Each of the other networks up the hierarchy consists of an abstracted version of its subordinate network. Each of the decision networks up the hierarchy corresponds to an increasingly coarser description of the world.

The structure of each abstract decision network is similar to that of the decision network used to model the real world. The root of the network is a hypothesis node whose values range over all possible maps corresponding to the degree of abstraction represented by that the decision network. The root node has a set of children, each one an abstract vertex, corresponding to a region (set of neighbor vertices) of the decision network immediately below in the hierarchy. The values for each of these range over all possible vertex configurations, just as they did in our original formulation. Each node corresponding to an abstract vertex has a set of children, corresponding to the abstract edges leaving that vertex. Each abstract edge represents the fact that there exists some edge in the lower, less abstract decision network, that allows the robot to travel between the two regions represented by the two endpoints of the abstract edge.

Figure 3 represents a sample hierarchy: it is made up of three rows, each one displaying a grid representing the world at some level of abstraction along with its corresponding decision network. At the top level, the world is represented by four connected regions $(1,1)_1, (1,2)_1, (2,1)_1, (2,2)_1$. Four abstract edges $a, b, c, d$ represent the connectivity between these four regions. At the next level in the hierarchy, we find a more detailed version of the world: Each of the four regions present in the previous grid has been expanded into four abstract regions of its own. In particular, note the two abstract edges $e, f$ connecting $(1,2)_2$ to $(1,3)_2$ and $(2,2)_2$ to $(2,3)_2$, respectively. The abstract edge $a$ linking regions $(1,1)_1, (1,2)_1$ exists iff one of $e, f$ exists as well. At the next level down, we simply have the grid and decision network representing the real world, as described in Section 3. Note that abstract edge $e$ depends on the presence of certain features in the real world (those that allow the robot to conclude one of two paths actually exist).

The computation of the possible values for each of the nodes in each of the decision trees in this hierarchy is straightforward, as is the computation of the conditional probabilities for each of the nodes. The same algorithms used to compute these on our original formulation can be used here.

One issue that arises in this approach is that of deciding when to switch to a more detailed level of reasoning, that is, a more precise decision model. An obvious alternative would be to make this switch when the robot has explored sufficiently to determine uniquely the value for the hypothesis node in the decision network currently in use. However, it is not clear that the number of possible values for the hypothesis node at the next, more detailed level will have been reduced sufficiently by then to allow the robot to evaluate that more detailed network. In fact, for large spaces this seems very unlikely. Another alternative would then be, to make the switch to the next more detailed level when it has shrunk to a manageable size. However, for large spaces it is not yet clear how much exploration this would require. Our current approach involves staying at a given level of abstraction until the complexity of the next more detailed level falls under a threshold value; this approach may still require that Huey adjust dynamically the set of maps under consideration for any given decision network, as mentioned in Section 3.

The actual decision model used for each of the levels in this hierarchy is similar to that of Section 3, modulo some differences introduced by the fact that the networks represent abstractions of the real world. The most important difference has to do with the definition of the cost functions used for the abstract decision networks. The cost function defined in Section 3 was defined in terms of the traversal of actual edges between point locations (the LDPs in the world). The cost functions for the abstract decision network must take into account both the actual traversal of the edge and some estimate of the cost of traversal inside the abstract regions represent by the nodes in the network. Additionally, in the early stages of exploration, we cannot assume that a path is know for each pair of locations in the graph. In this case, the cost of completing a task depends not only on the world, but also on how the robot gets to the destination. That is, there must be some method the robot uses for finding locations it does not have a path for, and this method gives rise to a cost function for tasks. The decision to be made is which method should be used. Presumably, some methods will rely as much as possible on known edges in an attempt to reach the destination quickly, but will add a minimal amount of new information. Other methods may be biased toward exploration, but take longer to complete a given task. We are currently considering various methods, and describe one of the former kind here.

The approach uses a weighting scheme on edges in the grid. We assign a weight of 1 to every edge known to exist and a weight of 0 to every edge known not to exist. For edges whose existence is not known, we can use the maps in the hypothesis node, $H$, to generate intermediate values as follows: for each edge, e, let $m$ be the number of maps in which e exists. The weight for e is then $\frac{m+1}{|H|+1}$. This scheme allow us to compute a value for any path in the grid by taking the product of the edges which make up the path. A path consisting of all known edges has value 1, while a path which includes an edge known to be non-existent has value 0. Given a task to reach a particular location, the robot takes a step along the path with the highest value, with shortest path length used to break ties. Any information gained by taking this step is included in the robot's map, and this process is repeated until the destination is reached. Other methods might include simply trying some shortest path even when the territory it crosses is unknown, deliberately avoiding known territory while trying to reach the goal, and random exploration.

Each one of these methods will have some associated cost, but unlike the simple model of the previous section, the actual cost may not be known for some methods before hand. We therefore anticipate the need to develop estimation functions for the costs of using each method. Using these estimations, we can make reasonable decisions between various methods based on their value both in reaching the goal and in providing new information.

# 5  An Approach to Designing Robot Control Systems

Our approach to designing Huey's control system is outlined as follows. We begin by considering Huey's overall decision problem, determining an optimal decision procedure according to a precisely stated decision-theoretic criteria, neglecting computational costs. We use an influence diagram to represent the underlying decision model and define the optimal procedure in terms of evaluating this model.

Huey's overall decision problem involves several component problems associated with specific classes of events occurring in the environment. These component decision problems include what action to take when approached by an unexpected object in a corridor, what sensor action to take next when classifying a junction, and what path to take in combin-ing exploration and task execution. Each of these problems is recurrent.

Problems involving what sensor action to take in classification or what path to take in navigation are predictably recurrent. For instance, during classification each sensor action takes about thirty seconds to a minute, so the robot has that amount of time to decide what the next action should be if it wishes to avoid standing idle lost in computation. The frequency with which choices concerning what path to take occur is dependent on how long Huey takes to traverse the corridor on route to the next LDP. With the current mobile platform operating in the halls of the computer science department, moving between two consecutive LDPs takes about four minutes. The problem of deciding what to do when approached by an unexpected object occurs unpredictably, and the time between when the approaching object is detected and when the robot must react to avoid a collision is on the order of a few seconds.

By making various (in)dependence assumptions and eliminating noncritical variables from the overall complex decision problem, we are able to decompose the globally optimal decision problem into sets of simpler component decision problems. Each of the sets of component problems are solved by a separate module. The computations carried out by these modules are optimized using a variety of techniques to take advantage of the expected time available for decision making [Kanazawa and Dean, 1989]. The different decision procedures communicate by passing probability distributions back and forth. For instance, the module responsible for making decisions regarding exploration and the module responsible for classifying LDPs pass back and forth distributions regarding the junction types of LDPs.

Our justifications for making (in)dependence assumptions and eliminating variables are sometimes statistical (e.g., based on some type of sensitivity analysis), but, more often than not, we take the expedient of simply trying something out and seeing if performance is adversely affected. We have considered the compilation methods of Heckerman, Breese, and Horvitz [Heckerman et al., 1989], but, so far, the combinatorics have made their decision-theoretic compilation methods impractical for our application. In our current approach, we make use of fixed-time decision procedures, and attempt to minimize the time the robot spends standing about just thinking. We are also looking at the use of variable-time decision procedures and doing run-time allocation of processor time [Boddy and Dean, 1989], but Huey's task does not appear to profit sig-nificantly from such attempts at optimization.

# 6 Conclusions and Future Work

This paper emphasizes two decision processes: one responsible for reasoning about the uncertainty inherent in a completely unfamiliar environment (Section 4), and a second responsible for assessing the expected value of various exploratory actions after some initial exploration has been done. (Section 3). Both processes must deal with noisy and ambiguous sensor data.

Our use of influence diagrams and Bayesian decision theory was inspired by recent work on decision-theoretic control for visual interpretation and sensor placement [Cameron and Durrant-Whyte, 1988, Hager, 1988, Levitt et al., 1988]. The design of the geographer module was based on the work of Kuipers [Kuipers and Byun, 1988] and Levitt [Levitt et al., 1987] on learning maps of large-scale space, and our own extensions to handle uncertainty [Basye et al., 1989]. The design of the module responsible for coordinating exploration and errand running was based on an application of information value theory [Howard, 1965].

Given that the complexity of Huey's exploratory strategy is largely determined by the number of possible maps $|M|$, we are looking for properties of the environment other than its global spatial layout that provide useful information for path planning. We have also come to realize that for Huey's sensors and the type of environments the robot is designed for, the most critical tradeoffs involve the cost of LDP classification and *map registration*: determining the robot's position with regard to its global map. Map registration is carried out by the geographer module, and losing registration (i.e., getting lost) can be quite costly for Huey. In the current system, we assume that there exist some small number of globally distinctive places, landmarks, that Huey can use for map registration. By carefully classifying LDPs, Huey can avoid costly registration. We are working on a decision model for exploration that allows for tradeoffs involving map registration and LDP classification.

# References

[Basye et al., 1989] Kenneth Basye, Thomas Dean,

and Jeffrey Scott Vitter. Coping with uncertainty in map learning. In *Proceedings IJCAI 11*, pages 663–668. IJCAI, 1989.

[Boddy and Dean, 1989] Mark Boddy and Thomas Dean. Solving time-dependent planning problems. In *Proceedings IJCAI 11*, pages 979–984. IJCAI, 1989.

[Cameron and Durrant-Whyte, 1988]
Alec Cameron and Hugh F. Durrant-Whyte. A bayesian approach to optimal sensor placement. Technical report, Oxford University Robotics Research Group, 1988.

[Dean *et al.*, 1990] Thomas Dean, Kenneth Basye, Robert Chekaluk, Seungseok Hyun, Moises Lejter, and Margaret Randazza. Coping with uncertainty in a control system for navigation and exploration. In *Proceedings AAAI-90*. AAAI, 1990.

[Hager, 1988] Gregory D. Hager. Active reduction of uncertainty in multi-sensor systems. Ph.D. Thesis, University of Pennsylvania, Department of Computer and Information Science, 1988.

[Heckerman *et al.*, 1989]
David E. Heckerman, John S. Breese, and Eric J. Horvitz. The compilation of decision models. In *UW89*, pages 162–173, 1989.

[Howard, 1965] Ronald A. Howard. Information value theory. *IEEE Transactions on Systems, Man, and Cybernetics*, 2:22–26, 1965.

[Jensen, 1989] Finn V. Jensen. Bayesian updating in recursive graphical models by local computations. Technical Report R 89-15, Institute for Electronic Systems, Department of Mathematics and Computer Science, University of Aalborg, 1989.

[Kanazawa and Dean, 1989] Keiji Kanazawa and Thomas Dean. A model for projection and action. In *Proceedings IJCAI 11*, pages 985–990. IJCAI, 1989.

[Kuipers and Byun, 1988] Benjamin J. Kuipers and Yung-Tai Byun. A robust, qualitative method for robot spatial reasoning. In *Proceedings AAAI-88*, pages 774–779. AAAI, 1988.

[Kuipers, 1978] Benjamin Kuipers. Modeling spatial knowledge. *Cognitive Science*, 2:129–153, 1978.

[Lauritzen and Spiegelhalter, 1988]
Stephen L. Lauritzen and David J. Spiegelhalter. Local computations with probabilities on graphical structures and their application to expert systems. *Journal of the Royal Statistical Society*, 50(2):157–194, 1988.

[Levitt *et al.*, 1987] Tod S. Levitt, Daryl T. Lawton, David M. Chelberg, and Philip C. Nelson. Qualitative landmark-based path planning and following. In *Proceedings AAAI-87*, pages 689–694. AAAI, 1987.

[Levitt *et al.*, 1988] Tod Levitt, Thomas Binford, Gil Ettinger, and Patrice Gelband. Utility-based control for computer vision. In *Proceedings of the 1988 Workshop on Uncertainty in Artificial Intelligence*, 1988.

[Pearl, 1988] Judea Pearl. *Probabilistic Reasoning in Intelligent Systems: Networks of Plausible Inference*. Morgan-Kaufman, Los Altos, California, 1988.

[Randazza, 1989] Margaret J. Randazza. The feature recognition module of the ldp system for the robot huey. M.Sc. Thesis, Brown University, 1989.